\documentclass[paper]{ite}                  
\usepackage{graphicx}
\usepackage[usenames]{color}
\usepackage{pbox}
\usepackage{times}
\usepackage{epsfig}
\usepackage{graphicx}
\usepackage{amsmath}
\usepackage{amssymb}
\usepackage[inline]{enumitem}
\usepackage{booktabs,colortbl,array}
\usepackage{latexsym}

\def\CNN{ConvNet}

\title{Visual Instance Retrieval with Deep Convolutional Networks}
\authorlist{%
 \authorentry{Ali S. Razavian}{n}{KTH}
 \authorentry{Josephine Sullivan}{n}{KTH}
 \authorentry{Stefan Carlsson}{n}{KTH}
 \authorentry{Atsuto Maki}{n}{KTH}
}
\affiliate[KTH]
 {Royal Institute of Technology (KTH)}
 {Stockholm, Sweden}

\acceptancePreIN{} 

\videoFlag{0} 

\begin{document}
\begin{abstract}
This paper provides an extensive study on the availability of 
image representations based on convolutional networks (ConvNets)
for the task of visual instance retrieval.
Besides the choice of convolutional layers, 
we present an efficient pipeline exploiting multi-scale 
schemes to extract local features, in particular, by taking geometric invariance 
into explicit account, i.e. positions, scales and spatial consistency.
In our experiments using five standard image retrieval datasets, 
we demonstrate that generic ConvNet image representations
can outperform other state-of-the-art methods if they are extracted appropriately. 
\end{abstract}


\maketitle

\section{Introduction}

Visual instance retrieval is amongst widely studied applications of computer vision, and the research progress has been reported based on several benchmark datasets. Using a database of reference images each having a label corresponding to the particular object or scene and given a new unlabeled query image, the task is to find images containing the same object or scene as in the query. Not to mention the accuracy of the retrieval results, there is also the challenge of search efficiency because modern image collections can contain millions if not billions of images. To solve the problem, one has two major design choices: \textit{image representations} and \textit{similarity measures} (or distance metrics). The task can be therefore viewed as encoding a large dictionary of image appearances where each visual instance needs a compact but descriptive representation.

 Image representations based on convolutional networks (ConvNets) are increasingly 
 permeating in many established application domains of computer vision.  
 Much recent work has illuminated
 the field on how to design and train ConvNets to maximize performance
 \cite{Simonyan:13,Girshick:14,bmvc:14:Chatfield,eccv:Zeiler:14,arxiv:v2:14:Azizpour}
 and also how to exploit learned ConvNet representations to solve
 visual recognition tasks \cite{Oquab14,Donahue14,corr:14:Fischer,cvpr:14:Razavian} 
 in the sense of transfer learning. 
 Based on these findings, although less explored, recent works also suggest the potential 
 of ConvNet representations for the task of image retrieval 
 \cite{eccv:14:Babenko,multimedia:14:Wan,cvpr:14:Razavian}, which we concern ourselves in this paper.

Possible alternatives for such a representation include holistic representations where the whole image is mapped to a single vector 
 \cite{eccv:14:Babenko,multimedia:14:Wan}  and multi-scale schemes where local features are extracted at multiple scale levels \cite{eccv:Gong:14,cvpr:14:Razavian}.
An essential requirement is that a proper representation should be ideally robust against variations such as scale and position of items in the image. That is, representations for the same instance in different viewpoints, or in various sizes, should be as close to each other as possible. Considering the geometric invariance as our priority, in this paper, we study the multi-scale scheme and design our pipeline including the similarity measure accordingly.

\begin{figure}[t]
  \begin{center}
    \begin{tabular}{cc}
    \includegraphics[width=\linewidth]{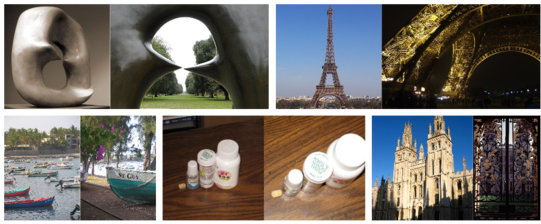} 
    \end{tabular}
    \end{center}
  \caption[]{\small In visual instance retrieval, items of interest often appear in different viewpoints, lightings, scales and locations. The primary objective of the paper is to introduce a generic pipeline that can well handle all those variations.}
  \label{fig:banner}
\end{figure}
Another practical aspect to consider for instance retrieval is the dimensionality and
memory requirements of the image representation. Usually, two separate
categories are considered; 
the \textit{small} footprint representations are encoding each image with less than 1kbytes 
and 
the \textit{medium} footprint representations which have dimensionality
between 10k and 100k. The former is required when the number of
images is huge and memory is a bottleneck, while the latter
is more useful when the number of images is less than 50k.
To the best of our knowledge,\textcolor{black}{this work is the first to show that the ConvNet representations are well suited for the task of both small and medium footprint retrieval while our main focus will be on the 
\textit{medium} footprint representations.}

The overall objective of this work is 
to highlight the potential of ConvNets image representations
for the task of visual instance retrieval 
while taking into account above mentioned factors, i.e. geometric invariance, the choice of layers, and search efficiency. 
In sum, the contributions of the paper are three-fold:
\begin{itemize}
\item We demonstrate that a ConvNet image representation can outperform state-of-the-art methods on all the five standard retrieval datasets if they are extracted suitably.  These datasets cover the whole spectrum of patterns from texture-less to repeated ones.

\item We formulate a powerful pipeline for retrieval tasks
  by extracting local features based on ConvNet representations.

\item We suggest and show that spatial pooling is well suited to the task of instance retrieval regarding dimensionality reduction and preserving spatial consistency.

\end{itemize}
It should also be noted that our pipeline does not rely on the bias of the dataset \textcolor{black}{as it heavily relies on the generic ConvNet train on ImageNet} (except
  to estimate the distribution of the data for the 
  representation). Therefore, it can be
  highly parallelized.
  For a fair comparison, the benchmarking of our methods will be with 
  pipelines that do not include any search refinement procedures such
  as query expansion \cite{iccv:07:Chum} and spatial
  re-ranking \cite{Philbin:cvpr:07,cvpr:09:Perdoch}.

\section{Related work}

It was not until Krizhevsky and Hinton
\cite{KrizhevskyH11} evaluated their binary codes that representations produced by deep learning were considered for image retrieval. Since the recent success of deep convolutional networks on image classification \cite{Krizhevsky12}, trained with large scale datasets \cite{imagenet_cvpr09}, the learned representations of ConvNet attracted increasing attention also for other applications of visual recognition. See for example \cite{Oquab14,eccv:Zeiler:14,Donahue14,arxiv:v2:14:Azizpour}. 
Among those a few papers have recently shown promising early results for image retrieval tasks
\cite{multimedia:14:Wan,cvpr:14:Razavian,eccv:14:Babenko,arxiv:Ng:15} using
ConvNet representations although the subject is yet to be explored. 


The key factor for the task of visual instance retrieval is to find an efficient image representation.
Popular approaches for this goal include vector aggregation methods such as the bag-of-words (BOW) representation \cite{iccv:03:Sivic}, the Fisher Vector (FV) \cite{cvpr:07:Perronnin}, and the VLAD representation \cite{pami:12:Jegou}.
The performance of \cite{pami:14:Simonyan} and \cite{Tolias:iccv:13} have been among the state-of-the-art in this domain.

Another line of work in retrieval deals with the compromise between performance and the representation size. The challenge of this task is to encode the most distinctive representation on small-footprint representations. Depending on the task, the size of representation can vary from 1k dimension to less than 16 bytes \textcolor{black}{\cite{cvpr:08:Torralba,pami:12:Jegou}}. 

The current paper is most related to 
\cite{multimedia:14:Wan,eccv:14:Babenko,cvpr:14:Razavian,eccv:Gong:14} concerning the usage of ConvNet representations. 
The work in \cite{eccv:14:Babenko,multimedia:14:Wan} for instance examined ConvNet representations taken from different layers of the network while
focusing on holistic features. 
The representation in the first fully connected layer was the choice in \cite{cvpr:14:Razavian}, and it was also concluded 
\cite{eccv:14:Babenko} to perform the best but still not better than other state-of-the-art holistic features (e.g. Fisher vectors \cite{cvpr:14:Jegou}). 
It is a natural next challenge to improve further the performance to the extent comparable to or better than the state-of-the-art approaches by taking different aspects of representations and similarity measures into consideration. 





%
\section{The ConvNet representation of an image}
\label{sec:convnet_rep}


We first review our design choices as regards to 
the ConvNet representation we obtain from a deep convolutional network, 
preceding the description of our 
general pipeline which is provided in the next section. 
%
%
%
%
%
The basic idea underlying our pipeline is 
that we have a way to extract a good feature representation of 
the input, $I$, whether it is an image or its sub-patch. 
Such a representation can be
defined by exploiting generic ConvNets trained on ImageNet.

\subsection{The choice of layer: the final convolution layer}
\label{sec:layer}

In the following, we employ the standard architecture for the convolutional network \cite{Krizhevsky12} consisting of multiple convolutional layers, \textcolor{black}{$Layer^j(I) (j=1,...,c)$}, and three fully connected layers. 
%



Now, let us visit possible interpretation on the function that each layer has \cite{eccv:Zeiler:14}: units in the first convolutional layer respond to simple patterns like edges or textures. Second layer's units aggregate those responses into more complex patterns. As the process continues further (deeper), units are expected to respond to more and more complex patterns in their corresponding receptive fields \cite{eccv:Zeiler:14,iclr:Zhou:15}. The first fully connected layer has pooled all the patterns together 
and the next two fully connected layers learn a non-linear classifier and estimate the probability of each class.
Nevertheless, 
the ability to recognize specific patterns should be an
integral component of an image retrieval system. 

%
Given this, it makes more sense to use the responses of {\it the last
convolutional layer} \textcolor{black}{($Layer^c(I)$)} each response in this layer corresponds to a
specific appearance pattern (up to the deformations allowed by the
max-pooling and activation function applied after each convolutional
layer) occurring in a specific large sub-patch of the original
image. 
In fact, experimental evidence is provided in the study for the factors of transferability \cite{arxiv:v2:14:Azizpour} that the last convolutional layer is the best alternative for instance retrieval. Our intuition is that, at this level, relevant information for instance retrieval that is not suited for classification is still preserved. 

Another benefit of using \textcolor{black}{$Layer^c(I)$} is that it enables us to extract local features for rectangular patches which is an important advantage for retrieval. This is in contrast to the case of using a fully connected layer where one would need to either crop a part of the image; this will cause some loss of information or break the aspect ratio, which is harmful to the task of visual retrieval.
With these motivations,
we use the response maps from \textcolor{black}{$Layer^c(I)$}, the last
convolutional layer in this paper. See Figure \ref{fig:size} for the empirical justifications.



\begin{figure}[t!]
  \begin{center}
    \begin{tabular}{cc}
    \includegraphics[width=.48\linewidth]{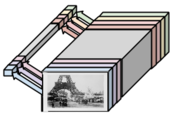} 
    \includegraphics[width=.48\linewidth]{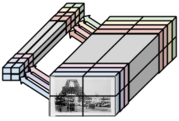} 
    \end{tabular}
    \end{center}
  \caption[]{\small \textbf{Schematic of spatial pooling.} 
    The sketches show how the dimensionality of the representation of
    a convolutional layer can be reduced by max-pooling 
	(left:1$\times$1 pooling, right: 2$\times$2 pooling).
        The width of the volumes corresponds to the number of filters
        used for convolutions.
        A 1$\times$1 pooling results in one number per feature map while 2$\times$2 pooling results in four numbers.
        See section \ref{sec:max-pooling} for the details. 
}
  \label{fig:spatial_pool_image}
\end{figure}
\subsection{Spatial max-pooling}
\label{sec:max-pooling}
%

As we use a deep and wide network, 
the dimensionality of this representation is also very large
(approximately 40$\times$40$\times$512).  It is neither feasible nor practical to
compute distances and similarities with vectors of this
size. Therefore some form of dimensionality reduction is necessary.

We reduce the dimension of our feature vector by exploiting the usual
max-pooling applied in the convolutional layers of a ConvNet. The difference is that we have a fixed number of cells in our spatial grid
as opposed to a spatial grid of cells of a fixed size. This type of pooling 
is not new and has been employed before. Because max-pooling as such allows one
to input images of variable size into the ConvNet and still extract a
fixed sized response before the first fully connected layer. 
Our experiments confirm that such a low-dimensional representation is more distinctive than the hand-crafted representation of the same memory footprint. Refer to section \ref{sec:result_small} for details.


If we apply spatial pooling with a grid of size 2$\times$2 to the original
(w$\times$h$\times$512) volume of response maps, we get a
2$\times$2$\times$512 (=2048) dimensional feature vector. Likewise, with 
a grid of size 1$\times$1, we get a 1$\times$1$\times$512 (= 512)
dimensional feature vector. See Figure \ref{fig:spatial_pool_image}. 
Max-pooling with a grid of size 2$\times$2 preserves \textcolor{black}{more spatial consistency than 1$\times$1}, which is useful for the task of retrieval. But max-pooling over a grid with too many cells generally tends to reduce performance.
In this paper, we also exploit the max-pooling to allow us to input larger
sized images to the ConvNet than were used while training the original
network. 
Spatial max-pooling is most useful when spatial consistency is the most distinctive feature.

\subsection{Image scale}
\label{sec:size}
\begin{figure}
  \begin{center}
    \begin{tabular}{cc}
      \includegraphics[width=.98\linewidth,trim=3cm 20cm 9cm 2cm]{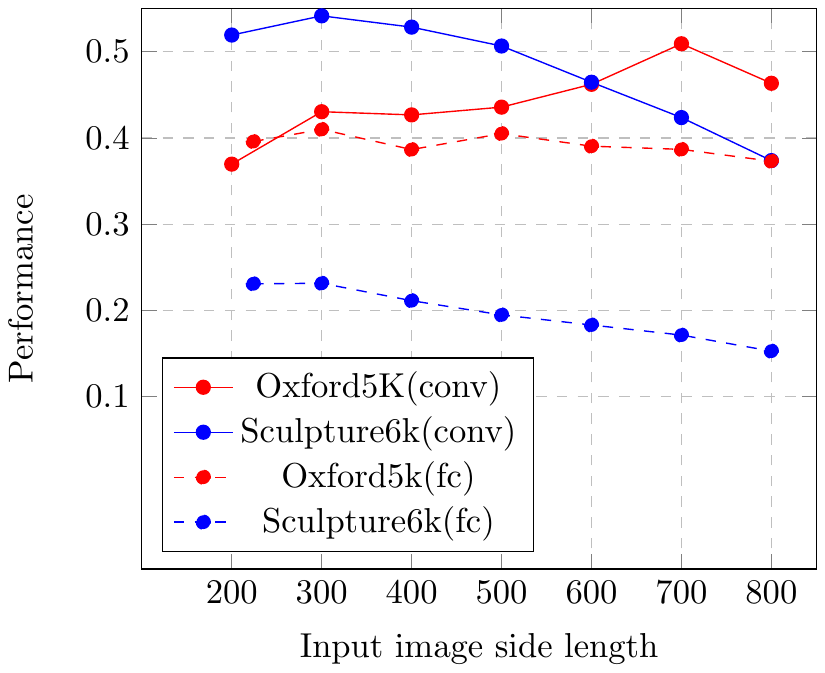} &
    \end{tabular}
    \end{center}
  \caption[]{\small \textbf{The effect of rescaling the input image and the choice of layer on retrieval task.} 
  Distinctive patterns for building landmarks are usually small details and by increasing the size of image, the performance on Oxford5k dataset increases while the most distinctive patterns for sculptures are the general shape of them and when the shape of sculpture becomes bigger than the size of the receptive fields, the performance starts to degrade. In our experiments, given the bias of datasets, we found that 600$\times$600 is a reasonable size for our target dataset.
  See section 5.1 for detailed discussion. \textcolor{black}{Also, Convolutional layers preserve more spatial information which is crucial to the task of retrieval. This observation solidifies the reported experiments of \cite{arxiv:v2:14:Azizpour}. Please refer to section 3.1 for more details.}}
  \label{fig:size}
\end{figure}
As discussed earlier, the units in convolutional layers respond to patterns in their receptive fields. Therefore, re-scaling the size of input image naturally results in a change in the activation of units in the response map. 
Figure \ref{fig:size} illustrates scaling effect on the performance followed by the explanation in section \ref{sec:result_params}.

 It should be 
noted that max-pooling operation needs be consistent across all the patches to provide meaningful comparisons and therefore all the images should be rescaled to have the same area.

\section{Pipeline for measuring similarities of images}
\label{sec:pipeline}

In this section, we describe our general pipeline exploiting multi-resolution search 
and approach 
%
to compute the distance (similarity) between a query and a reference image.

The items of interest (objects, buildings, etc.) in the reference and
query images can occur at any location and scale.
Then,  
different scales and other variations of input images 
may affect the behavior of convolutional layers as images pass through.
Facing with this potential issues, we choose to perform a form of 
multi-scale spatial search when we
compute the distance (similarity) between a query image and a reference image. 
Our search has two distinct sets of parameters and definitions: 
the first set describes the number, size and location of the sub-patches we
extract from each query and reference images $I$. The second set defines how we use the
sub-patches extracted from both the query image and the reference
image to compute the distance between the two images. We now describe
the two steps in more detail.



\begin{figure}[t]
  \begin{center}
      \includegraphics[width=.48\linewidth,trim=3cm 20cm 9cm 2cm]{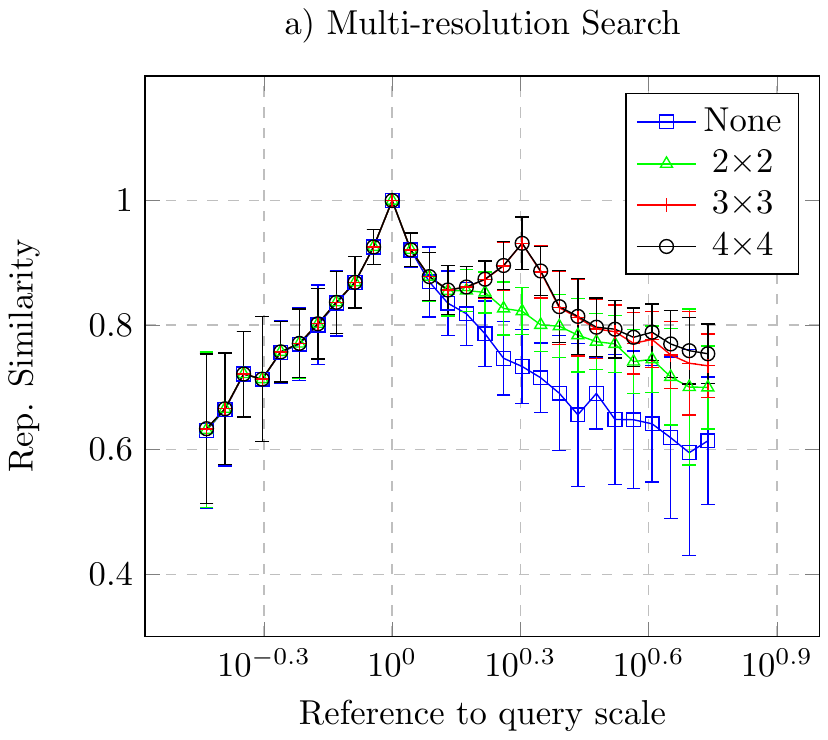}
      \includegraphics[width=.48\linewidth,trim=3cm 20cm 9cm 2cm]{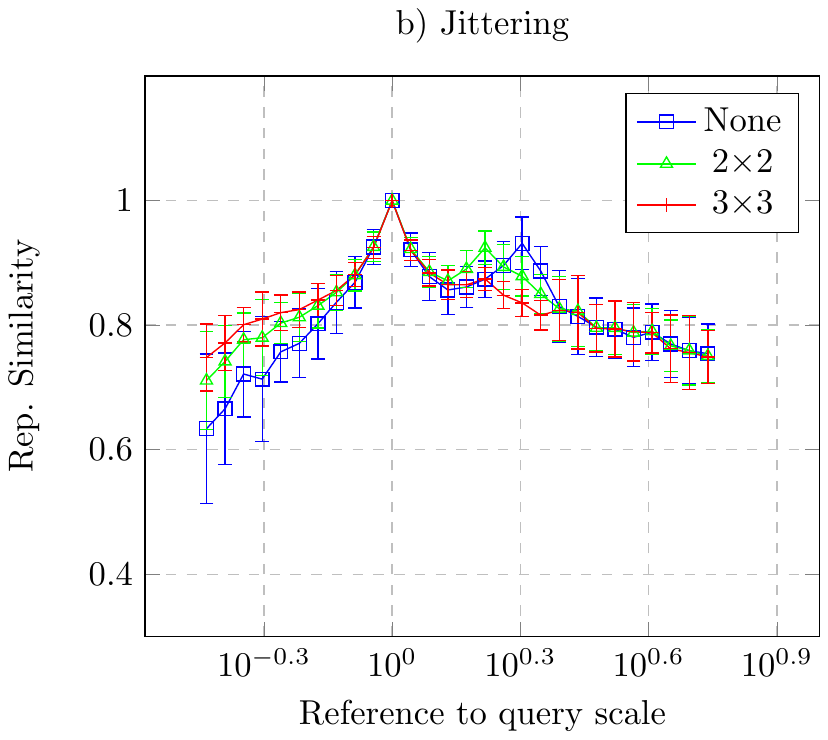}\\
      \includegraphics[width=.48\linewidth,trim=3cm 20cm 9cm 2cm]{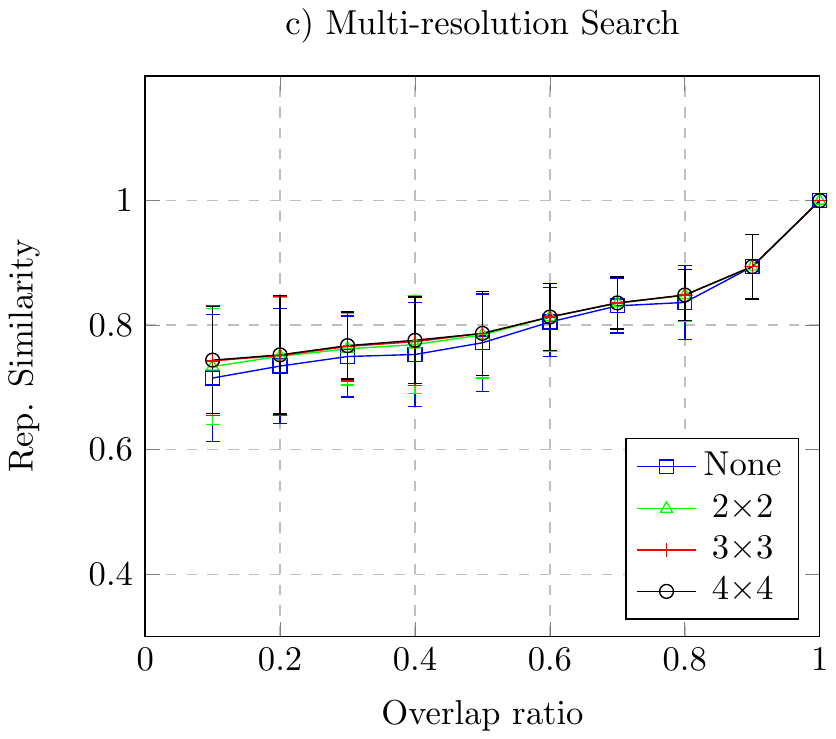}
      \includegraphics[width=.48\linewidth,trim=3cm 20cm 9cm 2cm]{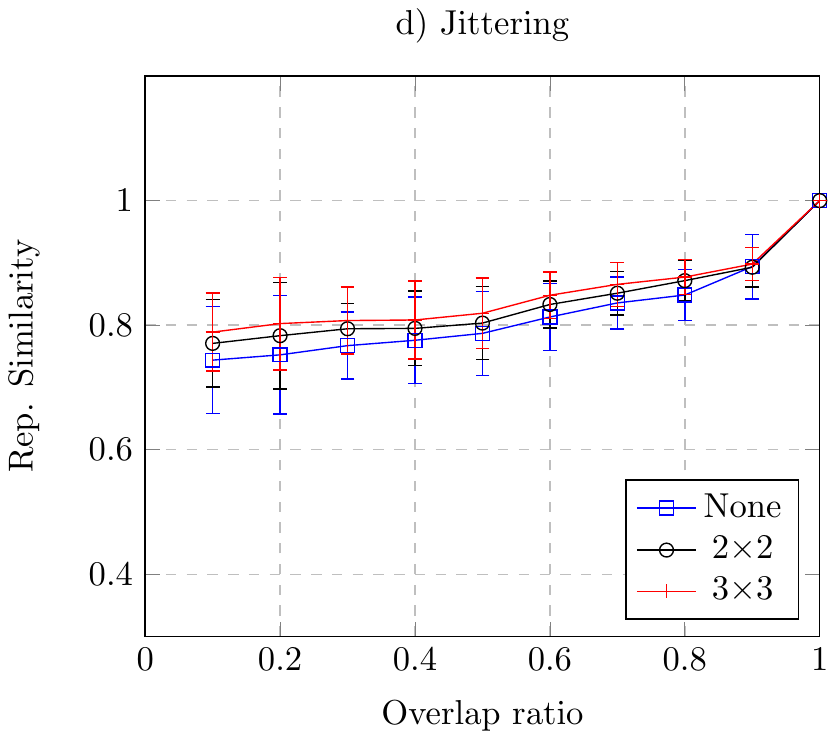}\\      
  \end{center}
  \caption[]{\textcolor{black}{\small \textbf{The effect of Multi-resolution search and Jittering on patch similarities:} a) Multi-resolution search is helpful when the item of interest has appeared on a smaller scale in the reference image than the query image. b) Jittering provides more robust results and is helpful in particular when the item of interest has appeared on a bigger scale in the reference image than the query. c \& d) Both Multi-resolution search and Jittering are helpful when the overlap between the query image and the reference image decreases. Please refer to \ref{sec:pipeline} for discussions.}}
  \label{fig:effects}
\end{figure}

\begin{figure}[t]
  \begin{center}
      \includegraphics[width=\linewidth]{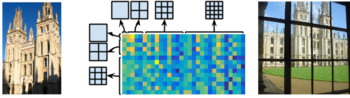}\\
    \end{center}
    \caption[]{\small \textbf{Schematic of distance matrix between two images:} The distance between two images is computed by pooling the distances between different patches of the query and the reference image according to equations (\ref{eq:patch_dist}-\ref{eq:qr_dist}).}
    \label{fig:multi_search}
\end{figure}

\subsection{Extracting local features}
\textbf{Multi-resolution search}
From each reference image, we extract multiple
sub-patches of different sizes at different locations. In fact,
we extract patches of $L$ different sizes. The different lengths of the side of the patches are
\begin{align}
  s=\text{max} \left(w_l = \frac{2 w}{l+1},   h_l = \frac{2 h}{l+1}\right), \quad\text{for $\;l = 1, \ldots, L$} 
\end{align}
For each square sub-patch of size $s$, we extract $l^2$
sub-patches centered at the locations
\begin{align}
 \left( \frac{w_l}{2} + (i-1)\,b_w, \frac{h_l}{2} + (j-1)\,b_h \right),\\
 \quad\text{where $\;b_w = \frac{w-w_l}{l-1}$, and~ $b_h = \frac{h-h_l}{l-1}$},(l \neq 1)\nonumber
\end{align}
for $i=1, \ldots, l$ and $j=1, \ldots, l$.  If the boundary of the square sub-patch exceeds the boundary of the image, we crop the rectangular sub-patch that is within the valid portion of the image.


Finally, we resize all the sub-patches to have the same area and then extract its ConvNet representation.
We should mention that we want all the sub-patches to have the same bias and in the ideal case, all the patches should have the same aspect ratio. To achieve this, we try to extract square sub-patches wherever possible and only extract rectangular sub-patches when there is no other option.

In our experiments, we set $L=4$ and therefore extract $30 (= \sum_{l=1}^4
l^2)$ sub-patches from each image. 

Multi-resolution search is most helpful when the item of interest has appeared on a smaller scale at an arbitrary position in the reference image.
For the justification on why extracting local patches from each image works for the task of instance retrieval, please refer to the recent work of \cite{cvpr:Tao:14}.

\textbf{Jittering}
 It is known that jittering increases the performance of \CNN{} representations \cite{Krizhevsky12}. That is, to have a more robust pipeline, instead of employing one patch from the query, we extract multiple patches in the same manner that we extracted sub-patches from reference images. 

Jittering is a powerful technique when the item of interest has appeared partially or in a bigger scale in the reference image.

\textcolor{black}{\textbf{Evaluation} To study the effectiveness of Multi-resolution search and Jittering, we evaluated these two methods on the similarity of 100 random patches extracted from Affine Covariant Regions Dataset \cite{ijcv:05:Mikolajczyk}. We isolated scaling from translation. See Figure \ref{fig:effects} for the evaluation results.}

%

\subsection{The distance between query and reference}

Next, we describe how we measure the distance
(similarity) between a query image $I^{(q)}$ and a reference image $I^{(r)}$.
The reference image $I^{(r)}$ generates sets of feature vectors $\quad\{\mathbf{f}^{(r)}_{i, j, l}\}$.
We define the distance between a
sub-patch, $I^{(q)}_*$, extracted from $I^{(q)}$ and the image $I^{(r)}$ as:
\begin{align}  
  d^*(I^{(q)}_{*}, I^{(r)}) = 
  \underset{1\leq r, s \leq m}{\underset{1\leq
    m \leq L,}{\min}}\; d\left(\mathbf{f}^{(q)}_{*}, \mathbf{f}^{(r)}_{r, s, m}
  \right) 
\label{eq:patch_dist}
\end{align}
where $d(\cdot, \cdot)$ corresponds to a distance measure between two
vectors. Typically $d(\cdot, \cdot)$ used in equation (\ref{eq:patch_dist}) is the $L_2$ normalized
distance.\footnote{
When $d(\cdot, \cdot)$ represents a similarity
measure, we replace the minimum in equation
(\ref{eq:patch_dist}) with a maximum.}

\begin{table}[t]
  \begin{center}
    {\scriptsize
    \begin{tabular}{@{}lc@{\hspace{1mm}}c@{\hspace{1mm}}c@{\hspace{1mm}}c@{\hspace{1mm}}c@{\hspace{1mm}}c@{}} 
	\toprule
    \textbf{Method} & \textcolor{black}{\#dim} & Oxford5k & Paris6k  & Holidays & Sculp6k & UKB\\
    \midrule
	Baseline & \textcolor{black}{512} &	46.2 &	67.4 & 74.6 & 46.5 & 90.6\\
    \midrule
 	MR$_{2\times2}$ &\textcolor{black}{2.5k} &	58.0	&	68.0	&	70.7	&	40.3	&	85.9\\
 	MR$_{3\times3}$&	\textcolor{black}{7k}&	63.4		&	68.5	&	70.2			&		39.8		&	77.5\\
 	MR$_{4\times4}$&	\textcolor{black}{15k}&	65.5		&	68.5	&	69.1			&		39.5		&	65.8\\
    \midrule
 	MR$_{4\times4}$ + Jtr$_{2\times2}$  &	\textcolor{black}{15k}&		71.6		&	76.3	&	78.5			&		40.1		&	88.0\\
 	MR$_{4\times4}$ + Jtr$_{3\times3}$  &	\textcolor{black}{15k}&		72.4		&	77.8	&	82.2			&		38.9		&	92.7\\
    \midrule
 	MR$_{4\times4}$ + PCAw &	\textcolor{black}{8k}&	73.0		&	71.0	&	74.7			&		36.1		&	84.2\\    
 	MR$_{4\times4}$ + Jtr$_{2\times2}$ + PCAw &	\textcolor{black}{8k}&	80.7		&	82.3	&	85.6			&		43.2		&	93.8\\
 	MR$_{4\times4}$ + Jtr$_{3\times3}$ + PCAw &	\textcolor{black}{8k}&	82.4		& 86.0	&	89.5			&		46.3		&	\textbf{95.8}\\
    \midrule
 	\pbox{20cm}{MR$_{4\times4}$ + Jtr$_{3\times3}$ + \\ SP$_{2\times2}$ + PCAw} &	\textcolor{black}{32k}&	\textbf{84.3}		&	\textbf{87.9}	&	\textbf{89.6}			&		\textbf{59.3}		&	95.1\\
    \midrule
BoB { \cite{Arandjelovic:iccv:11}}&& -&-&-&45.4&-\\

CVLAD { \cite{bmvc:13:Zhao}}&&51.4&-&82.7&-&90.5\\

PR-proj  { \cite{pami:14:Simonyan}}&&82.5&81.0&-&-&-\\

ASMK+MA { \cite{Tolias:iccv:13}}&& 83.8&80.5&88.0&-&-\\

    \bottomrule
    \end{tabular}
    }
  \end{center}
  \caption{\small \textbf{The effect of individual components in our pipeline}.
		\textbf{Baseline} refers to computing a single representation with $1\times1$ spatial pooling. 
\textbf{MR} refers to Multi-resolution search on the reference image where MR$_{2\times2}$ stands for the settings of $L=2$, and likewise \textbf{Jtr} refers to Jittering on the query image.
\textbf{PCAw} refers to PCA whitening and 
\textbf{SP} refers to spatial pooling.
}

    
  \label{tab:medium}
\end{table}


The final distance between
$I^{(q)}$ and $I^{(r)}$ is defined by the sum of each query
sub-patch's minimum distance to the reference image:
\begin{align}
  D(I^{(q)}, I^{(r)}) &=& \sum_{l=1}^L \sum_{i=1}^l \sum_{j=1}^l
   d^*(I^{(q)}_{i, j, l}, I^{(r)}) 
  \label{eq:qr_dist}
\end{align}
\textcolor{black}{Where $d^*$ is defined in equation \ref{eq:patch_dist}.}
See Figure \ref{fig:multi_search} for a schematic of computing the distance.
\subsection{Post-processing - PCA whitening}
There are some further optional, but standard post-processing steps that can be
applied after the max-pooling step and prior to computing the distance
between image-patches. First the representation is $L_2$
normalized. Then given a reference set of training images \textcolor{black}{($\textbf{X}$)} from the
dataset, a covariance matrix \textcolor{black}{$\textbf{C}$} is calculated \textcolor{black}{($\textbf{C}=\frac{(\textbf{X}-\mu_{X})^T(\textbf{X}-\mu_{X})}{n}$ where $n$ is the size of the reference set)} to allow PCA whitening of
the data such that a dimensionality reduction of the representation is enabled. 
We chose to reduce the dimensionality by half.

\section{Results}
We report results on five standard image retrieval datasets: \textbf{Oxford5k buildings} \cite{Philbin:cvpr:07}, 
 \textbf{Paris6k buildings} \cite{Philbin:cvpr:08}, \textbf{Sculptures6k} \cite{Arandjelovic:iccv:11}, \textbf{Holidays} \cite{jegou:eccv:08} and  \textbf{UKbench} \cite{nister:cvpr:06}.

\begin{table}[t]
  \begin{center}
    {\scriptsize
    \begin{tabular}{@{}lc@{\hspace{1mm}}c@{\hspace{1mm}}c@{\hspace{1mm}}c@{\hspace{1mm}}c@{\hspace{1mm}}c@{\hspace{1mm}}c@{}} 
    \toprule    
    \textbf{Method} & \#dim  & Oxford5k & Paris6k & Sculp6k & Holidays & UKB & Ox105k\\
	\midrule
    Max pooling + $l_1$ dist. & 256
    &53.3&\textbf{67.0}&\textbf{37.7}&71.6&84.2&48.9\\    
	Max pooling(quantized) +$l_2$ dist. & 32B &43.6&54.9&26.1&57.8&69.3&38.0\\
\midrule
VLAD+CSurf {\tiny \cite{multi:Xioufis:14}}&128 &  29.3&-&-& 73.8&83.0&-\\
mVLAD+Surf {\tiny \cite{multi:Xioufis:14}}&128 & 38.7&-&-& 71.8&87.5&-\\
T-embedding {\tiny \cite{cvpr:14:Jegou}}& 128 & 43.3&-&-& 61.7&85.0&35.3\\
T-embedding {\tiny \cite{cvpr:14:Jegou}}& 256 &  47.2&-&-& 65.7&86.3&40.8\\
	\midrule	
	Sum pooling + PCAw {\tiny \cite{babenko:iccv:15}} & 256 & 58.9 &-&-&80.2&\textbf{91.3}&\textbf{57.8}\\
	Ng et al, \cite{arxiv:Ng:15} & 128 &\textbf{59.3}&59.0&-&\textbf{83.6}&-&-\\
    \bottomrule
    \end{tabular}
    }
  \end{center} 
  \caption{\small \textbf{Performance of small memory
      footprint regimes.} The top two rows show our results. ConvNet based methods constantly
      outperform previous s.o.a. for small memory footprint
      representations. 
	  Two bottom results refer to the recent advancement in ConvNet based instance retrieval methods.
      For this experiment we trained our network similar to OxfordNet but with 256 kernels in the last convolutional layer.
       The result reported on Sculpture dataset was computed on $227\times227$ images.
      }    
  \label{table:small}
\end{table}

\label{sec:result_small}
\subsection{Medium footprint representation \& choice of parameters}
\label{sec:result_params}

To evaluate our model we used the publicly available network of \cite{corr:14:Simonyan14}. This network contains 16 convolutional layers followed by three fully connected layers. 
We should note that we extended the bounding box of query images by half the size of the receptive fields so that the center of receptive fields for all the units remains inside the bounding box. 

We have earlier displayed in Figure \ref{fig:size} 
the effect of resizing the input image size on the final result on two
datasets. The most interesting jump in performance due to increasing
the size of the input image is seen for the Oxford5k building
dataset. An explanation for this is as follows. The distinctive
elements of the images in the Oxford5k building dataset correspond to
relatively small sub-patches in the original image.  A response in 
the last convolutional layer, $L^c(I)$, 
corresponds to a whether a particular pattern is
present in a particular sub-patch of the input image. Increasing the
size of the input image means that the area of this sub-patch is a
smaller proportion of the area of the input image, i.e. 
the representation is
better at describing local structures. In contrast, increasing the 
resizing factor on the input image decreases the performance 
on the Sculpture6k dataset
because for these images better recognition seems to be achieved 
by describing the global structure. 
For all the experiments, we resized the images to have an area of 600$\times$600 pixels 
as it turns out to be a good dimension given these datasets.

In Table \ref{tab:medium} we
report results for our medium sized representations and compare their
performance to that of other s.o.a. image representations. From
examining the numbers one can see that our image retrieval
pipeline based on ConvNet representations outperform hand-crafted
image representations, such as VLAD and IFV, which mostly require additional
iterations of learning from specialized training data. 


\begin{figure}[t]
  \begin{center}
      \includegraphics[width=\linewidth]{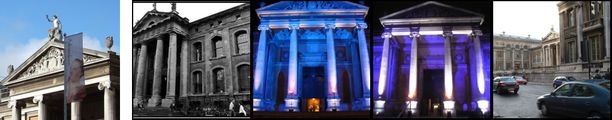}\\
      \includegraphics[width=\linewidth]{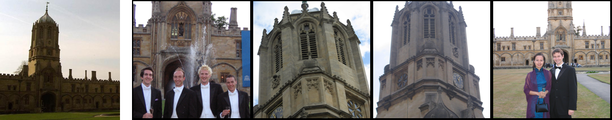}\\
      \includegraphics[width=\linewidth]{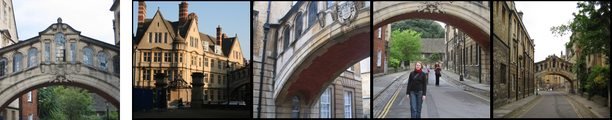}\\      
      \includegraphics[width=\linewidth]{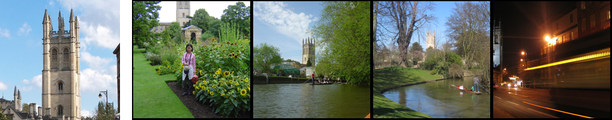}\\      
    \end{center}
    \caption[]{\small \textbf{ Visualization of relevant images with lowest score.} For query image (leftmost), four relevant images with the lowest similarity score have been retrieved. As expected, some portion of these images contain the item of interest with the most radical viewpoint or lighting changes. 	
        }
    \label{fig:oxford_failure}
\end{figure}

Table \ref{tab:medium} shows the results and the effect of each step in the pipeline.
\textbf{Baseline} refers to computing a single representation with $1\times1$ spatial pooling. It is evident that reported performance is relatively good already at this stage. 
Multi-resolution search (\textbf{MR}) helps the most in cases when the query item has appeared on a smaller scale in the reference image. The improvement is most noticeable in Oxford5k and Paris6k datasets where each query contains a full size landmark while in the reference set they have appeared in an arbitrary position and scale. On the other hand, employing \textbf{MR} alone degrades the performance on UKB dataset where items have appeared more or less at a uniform scale. 
Jittering (\textbf{Jtr}) and PCA whitening (\textbf{PCAw}) are almost always useful. Spatial pooling (\textbf{SP}) preserves spatial consistency, and it is always helpful as anticipated when the spatial position of patterns are the most distinctive feature. For example, it boosts the performance for Sculpture dataset where the boundary of the sculpture is the primary source of information. 
As a whole, the ConvNet representation with the proposed pipeline
outperforms all the state-of-the-art on all datasets. 
None of the numbers reported in this table considers the effect of post-processing techniques such as query expansion or re-ranking which can be commonly applied to different techniques including ours.


 
\begin{figure}
  \begin{center}
      \includegraphics[width=\linewidth]{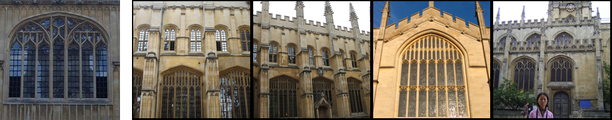}\\
      \includegraphics[width=\linewidth]{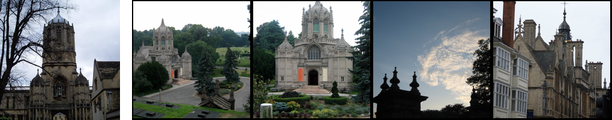}\\      
      \includegraphics[width=\linewidth]{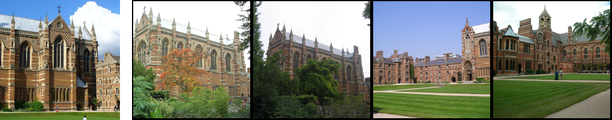}\\      
      \includegraphics[width=\linewidth]{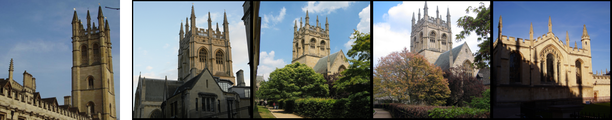}\\      
    \end{center}
    \caption[]{\small \textbf{ Visualization of irrelevant images with highest score.} For query image (leftmost), four irrelevant images with the highest similarity score have been retrieved. In some cases, it is even hard for humans to tell the difference apart.}
    \label{fig:oxford_false_pos}
\end{figure}

Figure \ref{fig:oxford_failure} shows the relevant images in reference set with lowest score for every query.
Figure \ref{fig:oxford_false_pos} displays irrelevant images with highest score to each query.

\subsection{Small footprint representations}
In Table \ref{table:small} we report results for our small 
memory representations and compare them to those by state-of-the-art methods with
128 or more dimensions. 
For the ConvNet methods, we can aggressively quantize the encoding of the numbers in each dimension of the representation to reduce the memory use.

Preceded by the initial introduction of our proposed pipeline in 2014 \cite{our_arxiv}, a few authors have advanced the performance of the small footprint retrieval pipelines with ConvNet. Ng et al. \cite{arxiv:Ng:15} replaced max pooling with traditional VLAD over the response map. Babenko and Lempitsky \cite{babenko:iccv:15} suggested that sum-pooling can perform better than max-pooling and Arandjelovic et al. \cite{15:Relja} included learning in the pipeline as opposed to the off-the-shelf usage. Note that those have been built on the basis of our pipeline which is formulated in the current paper.

\subsection{Memory and computational overhead}
The memory footprint and computational complexity of our pipeline is in the order of $O(L^3)$, and therefore, it is important to keep the $L$ small. In our experiments, we only extended $L$ to $4$ and the memory footprint corresponding to the last row of Table \ref{tab:medium} is $32k$ for reference images and $16k$ for query images.
The sub-patch distance matrix has around 120M elements for the task of Oxford5k and 350M elements for Holidays and the whole pipeline can fit in the RAM of an average computer.

Computing the distance matrix from sub-patch distance matrix according to equations (\ref{eq:patch_dist}-\ref{eq:qr_dist}) takes $30-40$ seconds on a \textit{single} CPU core and $50-60$ \textit{milliseconds} on a single K40 GPU in our experiment for the task of Oxford5k. This yields to a reasonable time complexity. All the modules in our pipeline can significantly benefit from parallelism.
To accelerate the feature extraction process, instead of extracting 30 representations for 30 sub-patches of an image, we feed the image in four different scales and extract the representation from response maps of the last Convolutional layer corresponding to the sub-patches as mentioned earlier.

\section{Conclusion}
We have presented an efficient pipeline for visual instance retrieval 
based on ConvNets image representations.
Our approach benefits from multi-scale scheme to extract local features
that take geometric invariance into explicit account.
For the representation, we chose to use last convolutional layer 
while adapting max-pooling, which not only made the layer available
in terms of its dimensionality but helped to boost the performance.
Throughout the experiments with five standard image retrieval datasets, 
we demonstrated that generic ConvNet image representations
can outperform other state-of-the-art methods in all the tested cases if they are extracted appropriately. 
Our pipeline does not rely on the bias of the dataset, and the only
recourse we make to specialized training data is in the PCA whitening.

For our future work, we plan to investigate domain adaptation to improve further the performance. For example, fine-tuning the ConvNet with landmark
dataset \cite{eccv:14:Babenko} increases the performance on Oxford5k up to \textbf{85.3}. We would like to highlight that our result should still be viewed as a baseline. Simple additions such as concatenating multi-scale, multi-layer and different architecture representations
give a boost in performance (e.g. \textbf{87.2} for Oxford5k).
From our measure of similarity, a rough bound can also be estimated within our framework, which may be useful for a novel search refinement techniques.

\end{document}